\documentclass[conference]{IEEEtran}
\usepackage{fancyhdr} 
\IEEEoverridecommandlockouts
\usepackage{cite}
\usepackage{amsmath,amssymb,amsfonts}
\usepackage[ruled,vlined]{algorithm2e}
\usepackage{graphicx}
\usepackage{textcomp}
\usepackage{xcolor}
\usepackage{subfig}
\usepackage{booktabs}
\usepackage{multirow}
\usepackage{lscape}
\usepackage{pifont}
\usepackage[pagebackref,breaklinks,colorlinks]{hyperref}

\usepackage{authblk}
\usepackage{blindtext}

\def\BibTeX{{\rm B\kern-.05em{\sc i\kern-.025em b}\kern-.08em
    T\kern-.1667em\lower.7ex\hbox{E}\kern-.125emX}}
\begin{document}

\title{SPATL: Salient Parameter Aggregation and Transfer Learning for Heterogeneous Federated Learning\\
}

\newcommand{\ali}[1]{{\color{purple}AA: #1}}

\newcommand{\proj}{\textsc{SPATL}\xspace}









\author[1]{Sixing Yu}
\author[1]{Phuong Nguyen}
\author[1]{Waqwoya Abebe}
\author[1]{Wei Qian}
\author[2]{Ali Anwar}
\author[1]{Ali Jannesari}
\affil[1]{Department of Computer Science, Iowa State University, Ames, IA, USA}
\affil[2]{IBM Almaden Research Center, San Jose, CA, USA}
\affil[1]{\textit {\{yusx, dphuong, wmabebe, wqi, jannesari\}@iastate.edu}}
\affil[2]{\textit {ali.anwar2@ibm.com}}

\maketitle

\thispagestyle{fancy}
\lhead{}
\rhead{}
\chead{}
\lfoot{\footnotesize{
SC22, November 13-18, 2022, Dallas, Texas, USA
\newline 978-1-6654-5444-5/22/\$31.00 \copyright 2022 IEEE}}
\rfoot{}
\cfoot{}
\renewcommand{\headrulewidth}{0pt}
\renewcommand{\footrulewidth}{0pt}

\begin{abstract}

Federated learning~(FL) facilitates the training and deploying AI models on edge devices. Preserving user data privacy in FL introduces several challenges, including expensive communication costs, limited resources, and data heterogeneity. In this paper, we propose SPATL, an FL method that addresses these issues by: (a) introducing a salient parameter selection agent and communicating selected parameters only; (b) splitting a model into a shared encoder and a local predictor, and transferring its knowledge to heterogeneous clients via the locally customized predictor. 
Additionally, we leverage a gradient control mechanism to further speed up model convergence and increase robustness of training processes.
Experiments demonstrate that SPATL reduces communication overhead, accelerates model inference, and enables stable training processes with better results compared to state-of-the-art methods. Our approach reduces communication cost by up to $86.45\%$, accelerates local inference by reducing up to $39.7\%$ FLOPs on VGG-11, and requires $7.4 \times$ less communication overhead when training ResNet-20.~\footnote{Code is available at: https://github.com/yusx-swapp/SPATL}

\end{abstract}


\section{Introduction}
\label{sec:intro}


Distributed machine learning~(ML) is extensively used to solve real-world problems in high performance computing~(HPC) environments. Typically, training data is first collected at a central location like a data center or HPC cluster. Afterwards, the data is carefully distributed across the cluster nodes based on the availability of resources. Training is conducted in a distributed manner and a resource and data aware fashion. However, new legislation such as General Data Protection Regulation~(GDPR)~\cite{gdpr} and Health Insurance Portability and Accountability Act~(HIPAA)~\cite{hipaa} prohibit user data collection.  In response to user privacy concerns, federated learning~(FL)~\cite{mcmahan2017fedavg} was proposed to train ML models while maintaining local data privacy~(restricting direct access to private user data). 





FL trains a shared model on edge devices (e.g., mobile phones) by aggregating locally trained models on a cloud/central server.
This setting, however, presents three key challenges:
First, the imbalance/non-independent identically distributed~(non-IID) local data easily causes training failure in the decentralized environment. 
Second, frequent sharing of model weights between edge devices and the server incurs excessive communication overhead.
Lastly, the increasing demand of computing, memory, and storage for AI models~(e.g., deep neural networks -- DNNs) makes it hard to deploy on resource-limited edge devices.
These challenges suggest that designing efficient FL models and deploying them effectively will be critical in achieving higher performance on future systems.
Recent works in FL and its variants~\cite{karimireddy2020scaffold,wang2020fednova,li2020fedprox} predominantly focus on learning efficiency, i.e., improving training stability and using the minimum training rounds to reach the target accuracy. However, these solutions further induce extra communication costs. As such, there is no superior solution to address the three key issues jointly. Furthermore, the above methods aim to learn a uniform shared model for all the heterogeneous clients. But this provides no guarantees of model performance on every non-IID local data.



Deep learning models are generally over-parameterized and can easily overfit during local FL updates; in which case, only a subset of salient parameters decide the final prediction outputs. It is therefore unnecessary to aggregate all the parameters of the model. 
Additionally, existing work~\cite{torrey2010translearn,weiss2016survey_transfer} demonstrate that a well-trained deep learning model can be easily transferred to non-IID datasets. 
Therefore, we propose to use transfer learning to address the data heterogeneity issue of Federated Learning.
As such, we train a shared model and transfer its knowledge to heterogeneous clients by keeping its output layers customized on each client.
For instance, computer vision models~(e.g., CNNs) usually consist of an encoder part~(embed the input instance) and a predictor head~(output layers).
In this case, we only share the encoder part in the FL communication process and transfer the encoder's knowledge to local Non-IID data using a customized local predictor.


Although we use the encoder-predictor based model as an example, our idea can be extend to all AI models whose knowledge is transferable~(i.e., we can transfer deep learning model by keeping the output layers heterogeneous on local clients).

Based on these observations, we propose an efficient FL method through \textbf{S}alient \textbf{P}arameter \textbf{A}ggregation and \textbf{T}ransfer \textbf{L}earning~({\proj}). Specifically, we train the model's encoder in a distributed manner through federated learning and transfer its knowledge to each heterogeneous client via locally deployed predictor heads. Additionally, we deploy a pre-trained local salient parameter selection agent to select the encoder's salient parameters based on its topology. Then, we customize the pre-trained agent on each local client by slightly fine-tuning its weights through online reinforcement learning. 
We reduce communication overhead by only uploading the selected salient parameters to the aggregating server.
Finally, we leverage a gradient control mechanism to correct the encoder's gradient heterogeneity and guide the gradient towards a generic global direction that suits all clients. This further stabilizes the training process and speeds up model convergence.
 



In summary, the contributions of {\proj} are:
\begin{itemize}
    \item {\proj} reduces communication overhead in federated learning by introducing salient parameter selection and aggregation for over-parameterized models. This also results in accelerating the model's local inference.
    \item \proj  addresses data heterogeneity in federated learning via knowledge transfer of the trained model to heterogeneous clients. 
    \item \proj utilizes a salient parameter selection agent by leveraging online reinforcement learning for fine-tuning.
    \item \proj enables scalable federated learning to allow large-scale decentralized training.
    \item We evaluate \proj on a medium-scale, 100 clients setup using Non-IID Benchmark~\cite{li2022NonIIDBench}. Our results show that compared to state-of-the-art FL solutions, when optimizing a model, \proj reduces communication cost by up to $7.4 \times$, improves model performance by up to 19.86\%, and reduces inference time by up to 39.7\% FLOPs.
\end{itemize}

%

\section{Motivation – Use Cases}
\label{sec:motivation}





With advancements in the performance of mobile and embedded devices, more and more applications are moving to decentralized learning on the edge. 
Improved ML models and advanced weight pruning techniques mean a significant amount of future ML workload will come from decentralized training and inference on edge devices~\cite{park2018dl_inference_fb}.
Edge devices operate under strict performance, power, and privacy constraints, which are affected by factors such as model size and accuracy, training and inference time, and privacy requirements. 
Many edge applications, such as self-driving cars, could not be developed and validated without HPC simulations, in which HPC accelerates data analysis and the design process of these systems to ensure safety and efficiency.

Therefore, the prevailing edge computing trend alongside FL requirements and edge constraints motivate \proj to address challenges in HPC.
Firstly, frequent sharing of model weights between edge devices and the central server incurs a hefty communication cost~\cite{problemsfl,Sheller2018medical}. Thus, reducing communication overhead is imperative. 
Secondly, the increasing demand for computing, memory, and storage for AI models (e.g., deep neural networks -- DNNs) makes it hard to deploy them on resource-constrained Internet-of-Things~(IoT) and edge devices~\cite{imteaj2022iot,nguyen2021iot}. Transfer learning can be a viable solution to address this problem.
Thirdly, latency-sensitive applications with privacy constraints (e.g., self-driving cars~\cite{kato2018auto}, augmented reality~\cite{mangiante2017vr}) in particular, are better suited for fast edge computing~\cite{wu2019ml_fb}. Hence, cutting back on inference time is quite important.
Tech giants like Google, Apple, and NVIDIA are already using FL for their applications (e.g., Google Keyboard~\cite{yang2018keyboard,chen2019keyboard}, Apple Siri~\cite{apple2019siri,paulik2021siri}, NVIDIA medical imaging~\cite{li2019nvidia}) thanks to their large number of edge devices. Hence, scalability is important in FL and HPC settings.
Lastly, training data on client edge devices depends on the user's unique usage causing an overall non-IID~\cite{mcmahan2017fedavg,zhao2018federated} user dataset. Data heterogeneity is a major problem in decentralized model training~\cite{problemsfl,nishio2019clientselection,zhao2018federated,hsieh2020quagmire,li2020convergence_noniid,lim2020federated,li2020fedprox,reddi2021fedopt,karimireddy2020scaffold,wang2020fednova,zhang2021adaptive_noniid}.
Thus designing efficient decentralized learning models and deploying them effectively will be crucial to improve performance of future edge computing and HPC.

\section{Related work}
\label{sec:relat}
\subsection{Federated Learning}
With increasing concerns over user data privacy, federated learning was proposed in~\cite{mcmahan2017fedavg}, to train a shared model in a distributed manner without direct access to private data. The algorithm FedAvg~\cite{mcmahan2017fedavg} is simple and quite robust in many practical settings. However, the local updates may lead to divergence due to heterogeneity in the network, as demonstrated in previous works~\cite{hsu2019measuring,karimireddy2020scaffold,li2020convergence_noniid}. To tackle these issues, numerous variants have been proposed~\cite{li2020fedprox,wang2020fednova,karimireddy2020scaffold}. For example, FedProx~\cite{li2020fedprox} adds a proximal term to the local loss, which helps restrict deviations between the current local model and the global model. FedNova~\cite{wang2020fednova} introduces weight modification to avoid gradient biases by normalizing and scaling the local updates. SCAFFOLD~\cite{karimireddy2020scaffold} corrects update direction by maintaining drift variates, which are used to estimate the overall update  direction of the server model. 
Nevertheless, these variants incur extra communication overhead to maintain stable training. Notably, in FedNova and SCAFFOLD, the average communication cost in each communication round is approximately $2 \times$ compared to FedAvg.

Numerous research papers have addressed data heterogeneity~(i.e. non-IID data among local clients) in FL ~\cite{zhao2018federated,hsieh2020quagmire,lim2020federated,zhang_fedufo,gong_ensemblefl,caldarola_graphfl,sun_soteria,horvath2021fjord}, such as adjusting classifier~\cite{luo2021no}, improving client sampling fairness~\cite{nishio2019clientselection},  adapting optimization~\cite{zhang2021adaptive_noniid,han2020adaptive_FLquantization,reddi2021fedopt,yu2021adaptive}, correcting local updates~\cite{karimireddy2020scaffold,li_model-contrastive,wang2020fednova}, using a tiering mechanism to synchronously update local model parameters within tiers and asynchronously update the global model~\cite{chai2021fedat}, generating client models from a central hypernetwork model~\cite{Shamsian2021pFedHN}, dynamically measuring local model divergence and adaptively adjusting to optimize hyper-parameters~\cite{zhuang2022fedema}, and using data-free knowledge distillation approach to address heterogeneous FL~\cite{zhu2021fedgen}.

Further more, federated learning has been extended in real life applications~\cite{liu_feddg,guo_multi-institutional}. One promising solution is personalized federated
learning~\cite{dinh2021personalized_moreau,huang2021personalized_crosssilo,zhang2021personalized_modelopt,fallah2020personalized_meta,hanzely2020lowerbound_personazlied,ozkara2021quped}, which tries to learn personalized local models among clients to address data heterogeneity. These works, however, fail to address the extra communication overhead.
However, very few works, such as~\cite{guo2020VeriFL_communication, wu2022knowledgeD_communication} focus on addressing communication overhead in FL. They either use knowledge distillation, or aggregation protocol,
the communication overhead reduction is not significant.

Additionally, benchmark federated learning settings have been introduced to better evaluate the FL algorithms. FL benchmark LEAF~\cite{caldas2019leaf} provides benchmark settings for learning in FL, with applications including federated learning, multi-task learning, meta-learning, and on-device learning.
Non-IID benchmark~\cite{li2022NonIIDBench} is an
experimental benchmark that provides us with Non-IID splitting of CIFAR-10 and standard implementation of SOTAs.
Framework Flower~\cite{beutel2020flower} provides FL SOTA baselines and is a collection of organized scripts used to reproduce results from well-known publications or benchmarks.
IBM Federated Learning~\cite{ibmfl2020ibm} provides a basic fabric for FL on which advanced features can be added. It is not dependent on any specific machine learning framework and supports different learning topologies, e.g., a shared aggregator and protocols. It is meant to provide a solid basis for federated learning that enables a large variety of federated learning models, topologies, learning models, etc., particularly in enterprise and hybrid-Cloud settings.

\subsection{Salient Parameter Selection}
Since modern AI models are typically over-parameterized, only a subset of parameters determine practical performance. 
Several network pruning methods have been proposed to address this issue.  These methods have achieved outstanding results and are proven techniques to drastically shrink model sizes. However, traditional pruning methods~\cite{gao_network_2021,liu_learnable_2021,wang_convolutional_2021} require time-consuming re-training and re-evaluating to produce a potential salient parameter selection policy. Recently, AutoML pruning algorithms~\cite{li2020eagleeye,chin2020legr} offered state-of-the-art~(SoTA) results with higher versatility. In particular, reinforcement learning~(RL)-based methods~\cite{yu2021gnnrl,yu2021gnnrl1,he2018amc,yu2021agmc}, which model the neural network as graphs and use GNN-based RL agent to search for pruning policy present impressive results.
However, AutoML methods need costly computation to train a smart agent, which is impractical to deploy on resource-limited edge FL devices. 

The enormous computational cost and effort of network pruning makes it difficult to directly apply in federated learning.
To overcome challenges of previous salient parameter selection methods and inspired by the RL-based AutoML pruning methods, we utilize a salient parameter selection RL agent pre-trained on the network pruning task. 
Then with minimal fine-tuning, we implemented an efficient salient parameter selector with negligible computational burden.




\begin{figure}
\begin{center}
  \includegraphics[width=.95\linewidth]{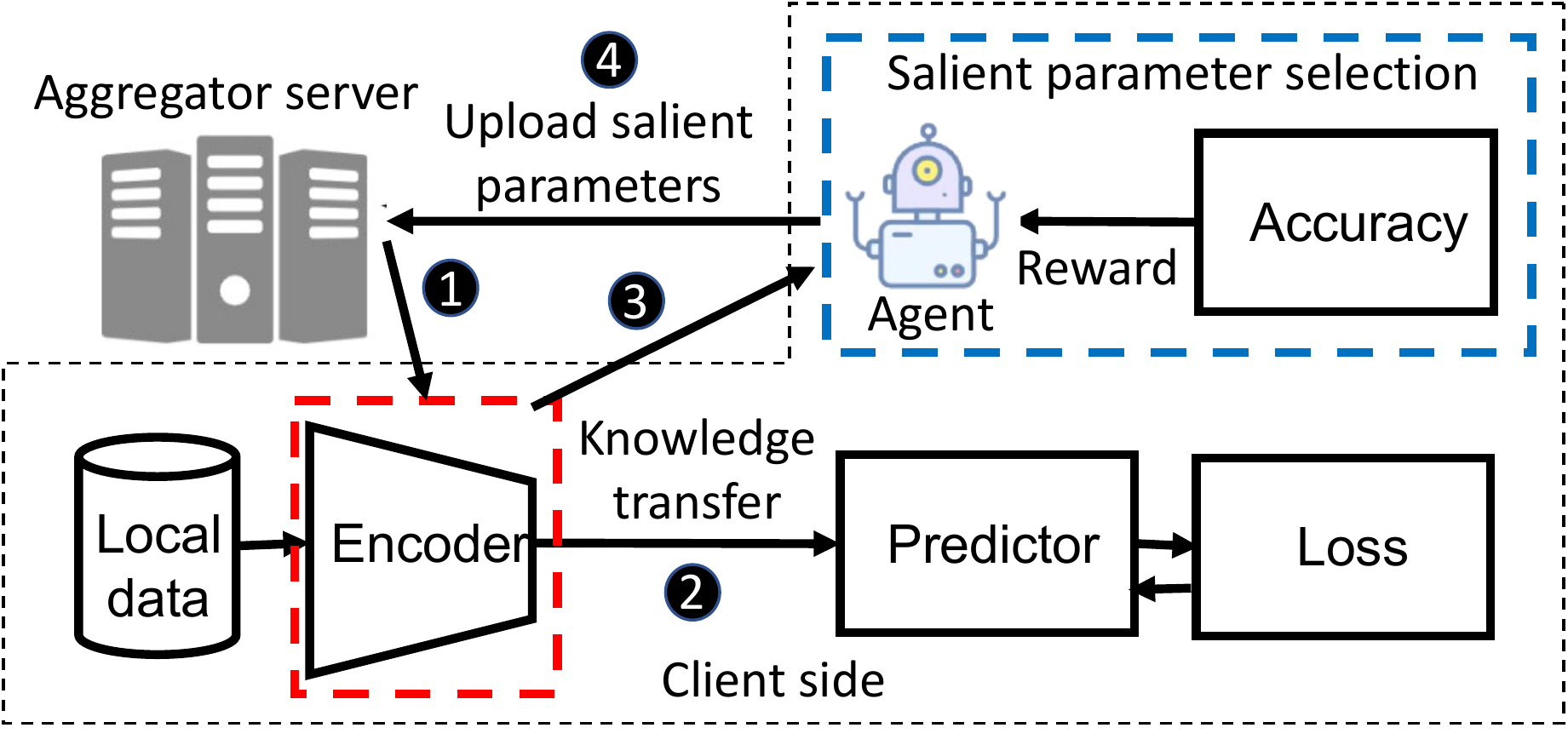}
\end{center}
   \caption{SPATL Overview. SPATL trains a shared encoder through federated learning, and transfers the  knowledge to heterogeneous clients. Clients upload salient parameters selected by a pre-trained RL-agent. The selected parameters are then aggregated by the server.}

\label{fig:overview}
\end{figure}

\section{Methodology}
\label{sec:metho}
\proj consists of three main components: knowledge transfer learning, salient parameter selection agent, and gradient control federated learning.  Figure~\ref{fig:overview} shows the \proj overview.
Unlike mainstream FL solutions, which attempt to train the entire deep learning model, \proj only trains the encoder part of the model in a distributed manner and transfers the knowledge to heterogeneous clients.   
In each round of federated learning, the client first downloads the encoder from the cloud aggregator~(\ding{202} in Figure~\ref{fig:overview}) and transfers its knowledge using a local predictor through local updates~(\ding{203} in Figure~\ref{fig:overview}). After local updates, the salient parameter selection agent will evaluate the training results of the current model based on the model performance~(\ding{204} in Figure~\ref{fig:overview}), and finally selected clients send the salient parameters to the server~(\ding{205} in Figure~\ref{fig:overview}). Additionally, both clients and the server maintain a gradient control variate to correct the heterogeneous gradients, in order to stabilize and smoothen the training process.  

\subsection{Heterogeneous Knowledge Transfer Learning}
Inspired by transfer learning~\cite{torrey2010translearn}, \proj aims to train an encoder in FL setting and address the heterogeneity issue through transferring the encoder's knowledge to heterogeneous clients.
Formally, we formulate our deep learning model as an encoder $E(w_e, x)$ and a predictor $P(w_p, e)$, where $w_e$ and $w_p$ are encoder and predictor parameters respectively, $x$ is an input instance to the encoder and $e$ is an input instance to the predictor (or embedding).



\proj shares the encoder $E(w_e, x)$ with the cloud aggregator, while the predictor $P^k(w^k_p, e)$ for the $k^{th}$ client is kept private on the client.
The forward propagation of the model in the local client $k$ is formulated as follows:
\begin{equation}
    e = E(w_e, x) ,
\end{equation}
\begin{equation}
    \hat{y} = P^k(w_p^k, e)
\end{equation}


\noindent During local updates, the selected $k^{th}$ client first downloads the shared encoder parameter, $w_e$, from the cloud server and optimizes it with the local predictor head, $w^k_p$, through back propagation. Equation~\ref{eq:loss1} shows the optimization function.
\begin{equation}
\label{eq:loss1}
    \min_{w_e, w^k_p} \mathcal{L}(w_e,w^k_p) = \frac{1}{n_i}\sum l(w_e,w^k_p,x_i,y_i)
\end{equation}
Here, $l$ refers to the loss when fitting the label $y_i$ for data $x_i$, and $n_i$ is the constant coefficient.

In federated learning, not all clients are involved in communication during each round. In fact, there is a possibility a client might never be selected for any communication round. Before deploying the trained encoder on such a client, the client will download the encoder from the aggregator and apply local updates to its local predictor only. After that, both encoder and predictor can be used for that client. Equation~\ref{eq:loss2} shows the optimization function.
\begin{equation}
\label{eq:loss2}
    \min_{w^k_p} \mathcal{L}(w^k_p) = \frac{1}{n_i}\sum l(w_e,w^k_p,x_i,y_i)
\end{equation}

\subsection{RL-based Topology-Aware Salient Parameter Selection}
One key issue of FL is the high communication overhead caused by the frequent sharing of parameters between clients and the cloud aggregator server. Additionally, we observed that deep learning models~(e.g., VGG~\cite{simonyan2015vgg} and ResNet~\cite{he2016resnet}) are usually bulky and over-parameterized. As such, only a subset of salient parameters decide the final output.
Therefore, in order to reduce the communication cost, we implemented a local salient parameter selection agent for selecting salient parameters for communication.

\begin{figure}
\begin{center}

\centerline{\includegraphics[width=.8\columnwidth]{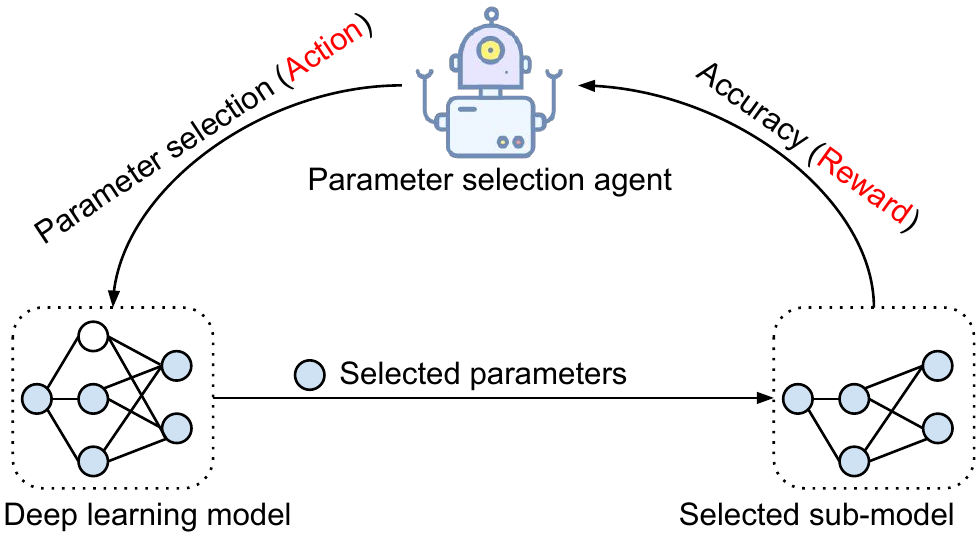}}
\caption{Parameter selection with reinforcement learning. The parameter selection agent selects a salient parameter and uses the selected sub-model's accuracy to search for the desired policy.}

\label{fig:pram_selction}
\end{center}
\end{figure}

Figure~\ref{fig:pram_selction} shows the idea of a salient parameter agent. 
Specifically, inspired by topology-aware network pruning task~\cite{yu2021agmc,yu2021gnnrl}, we model the neural network~(NN) as a simplified computational graph and use it to represent the NN's states. Since NNs are essentially computational graphs, their parameters and operations correspond to nodes and edges of the computational graph. 
We then introduced the graph neural network~ (GNN)-based reinforcement learning~(RL) agent, which takes the graph as input~(RL's environment states) and produces a parameter selection policy from the topology through GNN embedding. Additionally, the RL agent uses the selected sub-model's accuracy as reward to guide its search for the optimal pruning policy.
Training a smart agent directly through RL, however, is costly and impractical to deploy on the edge. 
To address this issue, we first pre-train the salient parameter agent in the network pruning task, and then customize the pre-trained agent on each local client by slightly fine-tuning its weights through online reinforcement learning~(detailed hyper-parameter setting in section~\ref{sec:eval}).

\SetKwInput{KwInput}{Input}
\SetKwInput{KwOutput}{Output}

\begin{algorithm}[t] {
\caption{Salient parameter selection with PPO}
\label{alg:ppo}
\KwInput{shared encoder $E(w_e)$ and model size constraints $d$} 

\KwOutput{selected salient parameter and corresponding index}

\For{search step $k = 1,2,..$}{
    $\hat{E}(\hat{w_e}) \leftarrow E({w_e})$
    
    $G \leftarrow$ computational graph of $\hat{E}(\hat{w_e})$
    
    \While{$Size(\hat{E}) > d$}{
        {$a = PPO.Actor\_Critic(G)$ }

        $PPO.Memory(G,a)$
        
        $\hat{E}(\hat{w_e}) \leftarrow$
        pram\_select($\hat{E}(\hat{w_e}),a$)
        
        $G \leftarrow$ computational graph of $\hat{E}$
    }
    $Reward = Accuracy(\hat{E})$
    
    \If{is best Reward} {
         $w \leftarrow \hat{w_e}$
         
         $idx \leftarrow$ extract the index of $\hat{w_e}$
    }
    $PPO.update(Reward)$
}
{\bfseries Return} $w, idx$
}
\end{algorithm}

\begin{figure}[t]
 \begin{center}

\centerline{\includegraphics[width=.5\columnwidth]{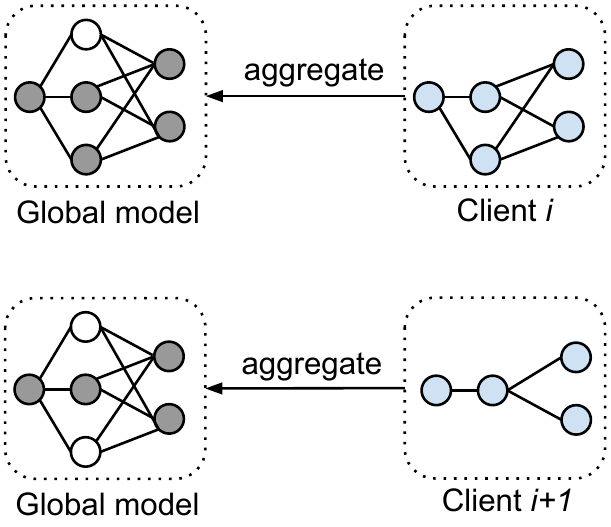}}
\caption{Aggregation with salient parameter. For each client, the aggregator will aggregate partial parameters corresponding to the client's salient parameter.}

\label{fig:pram_aggre}
\end{center}
\vspace{-10pt}
 \end{figure}

\subsubsection{Reinforcement Learning Task Definition}
Defining environment states, action space, reward function, and RL policy are essential for specifying an RL task. In this section, we will discuss these components in more detail. 
Algorithm~\ref{alg:ppo} shows the RL search process. 
For search step, we first initialize the target encoder $\hat{E}(\hat{w_e})$ with the input encoder ${E}({w_e})$, and convert it to a graph. If the size of $\hat{E}$ does not satisfy the constraints, the proximal policy optimization~(PPO)~\cite{schulman2017ppo} RL agent will produce a parameter selection policy $a$~(i.e., the action of the RL), to update $\hat{E}$. If $\hat{E}$ satisfies the size constraint, the RL agent will use its accuracy as reward to update the policy. Finally, the parameter $w$ and corresponding parameter index $idx$ of the target encoder $\hat{E}$ with the best reward will be uploaded to the cloud server. 

\noindent\textbf{Environment States.} 
We use a simplified computational graph $G(v,e)$ to represent the NN model~\cite{yu2021agmc}. In a computational graph, nodes represent hidden features~(feature maps), and edges represent primitive operations~(such as `add', `minus', and `product'). Since the NN model involves billions of operations, it's unrealistic to use primitive operations. Instead, we simplified the computational graph by replacing the primitive operations with machine learning operations~(e.g., conv 3x3, Relu, etc.).



\noindent\textbf{Action Space.} The actions are the sparsity ratios for encoder's hidden layers. The action space is defined as $a\in [0,1]^{N }$, where $N$ is the number of encoder's hidden layers.
The actor network in the RL agent projects the NN's computational graph to an action vector, as shown in equations~\ref{eq:graphencoder} and~\ref{eq:mlp}.
\begin{equation}
    g = GraphEncoder(G) ,
    \label{eq:graphencoder}
\end{equation}
\begin{equation}
    a = mlp(g) 
    \label{eq:mlp}
\end{equation}
Here, $G$ is the environment state, $g$ is the graph representation, and MLP is a multi-layer perceptron neural network. The graph encoder learns the topology embedding, and the MLP projects the embedding into hidden layers' sparsity ratios.

\noindent\textbf{Reward Function.} The reward function is the accuracy $\times 100$ of selected sub-network on validation dataset.

\begin{equation}
    Reward = Accuracy\times 100 
    \label{eq:reward}
\end{equation}



\subsubsection{Policy Updating}
The RL agent is updated end-to-end through the PPO algorithm. The RL agent trains on the local clients through continual online-learning over each FL round.
Equation~\ref{eq:1} shows the objective function we used for the PPO update policy. 
\begin{equation}
\label{eq:1}
    L(\theta) = \hat{\mathbb{E}_t}[min(r_t(\theta)\hat{A}_t, clip(r_t(\theta),1-\epsilon,1+\epsilon)\hat{A}_t)]
\end{equation}
Here, $\theta$ is the policy parameter~(the actor-critic network's parameter),
$\hat{\mathbb{E}_t}$ denotes the empirical expectation over time steps,
$r_t(\theta)$ is the ratio of the probability under the new and old policies, respectively,
$\hat{A}_t$ is the estimated advantage at time t, and $\epsilon$ is a clip hyper-parameter, usually 0.1 or 0.2.

\subsection{Generic Parameter Gradient Controlled Federated Learning}

Inspired by stochastic controlled averaging federated learning~\cite{karimireddy2020scaffold}, we propose a generic parameter gradient controlled federated learning to correct the heterogeneous gradient.
Due to client heterogeneity, local gradient update directions will move towards local optima and may diverge across all clients. To correct overall gradient divergence by estimating gradient update directions, we maintain control variates both on clients and the cloud aggregator. 
However, controlling the entire model's gradients will hurt the local model's performance on non-IID data. In order to compensate for performance loss, \proj only corrects the generic parameter's gradients~(i.e., the encoder's gradients) while maintaining a heterogeneous predictor. Specifically in equation~\ref{eq:control_variates}, during local updates of the encoder, we correct gradient drift by adding the estimate gradient difference $(c_g - c_l)$. 
\begin{equation}
    \label{eq:control_variates}
    w_e \leftarrow w_e - \eta \triangledown(\mathcal{L}(w_e,w_p;b) { + c_g - c_l}) 
\end{equation}
Here, control variate $c_g$ is the estimate of the global gradient direction maintained on the server side, and $c_l$ is the estimate of the update direction for local heterogeneous data maintained on each client.
In each round of communication, the $c_l$ is updated as equation~\ref{eq:cl_update}:
\begin{equation}
    \label{eq:cl_update}
c_l^* \leftarrow  c_l - c_g + \frac{1}{E\eta}(w_g - w_e)
\end{equation}
Here, $E$ is the number of local epochs, and $\eta$ is the local learning rate, while $c_g$ is updated by equation~\ref{eq:cg_update}:
\begin{equation}
    \label{eq:cg_update}
c_g \leftarrow  c_g + \frac{1}{|N|}\sum_{k \in K} \Delta c^k 
\end{equation}
Here, $\Delta c^k$ is the difference between new and old local control variates $c_l$ of client $k$, $N$ is the set of clients, and $K$ is the set of selected clients. 

Algorithm~\ref{alg:aggregate} shows \proj with gradient controlled FL.
In each update round, the client downloads the global encoder's parameter $w_g$ and update direction $c_g$ from server, and performs local updates. When updating the local encoder parameter $w_e$, $(c_g - c_l)$ is applied to correct the gradient drift. The predictor head's gradient remains heterogeneous.  Before uploading, the local control variate $c_l$ is updated by estimating the gradient drift.

\begin{algorithm}
    \caption{\proj with gradient controlled FL}
    \label{alg:aggregate}

{\bfseries Server executes:} 

initialize $w_g$, $c_g$.

\For{each round $t =1, 2,\ldots, T$} {
    
    $K \leftarrow$ random set of clients $\in N$
    
    \For{each client $k \in K$ {\bfseries in parallel}}{

        \textbf{communicate with client k}
        
        ${w}^k,i^k, {\Delta {c^k}}\leftarrow$ ClientUpdate$(w_g, {c_g})$
    }

    $w_g \leftarrow  w_g + \eta(\frac{1}{|K|} \sum_{k \in K} (w_g[i^k,:,:] - w^k)) $  
        
    {$c_g \leftarrow  c_g + \frac{1}{|N|}\sum_{k \in K} \Delta c^k $}
}

\;

{\bfseries ClientUpdate($w_g, {c_g}$):}

$\mathcal{B} \leftarrow$ split local dataset into batches 

initialize the local encoder $w_e \leftarrow w_g$, and control $c_l$

\For{epoch$=1,2,\ldots, E$} {
    \For{batch $b\in \mathcal{B}$}{

        $w_e \leftarrow w_e - \eta \triangledown(\mathcal{L}(w_e,w_p;b) { - c_l + c_g}) $ 
        
        $w_p \leftarrow w_p - \eta \triangledown(\mathcal{L}(w_e,w_p;b) $ 

    }
}

{$c_l^* \leftarrow  c_l - c_g + \frac{1}{E\eta}(w_g - w_e)$}

$\Delta c \leftarrow (c_l^* -c_l)$

$w_e, idx \leftarrow$ SalientPrameterSelection($E(w_e), d$)

\textbf{communicate} $w_e, idx, \Delta c$

{$c_l \leftarrow c_l^*$}

\end{algorithm}

\begin{figure*}
 \begin{center}

\centerline{\includegraphics[width=1.015\linewidth]{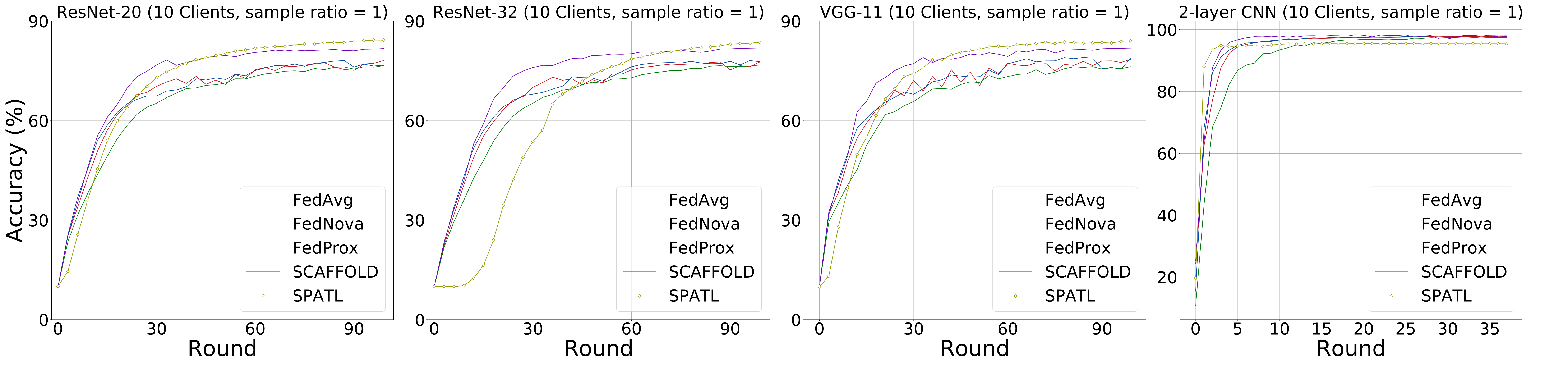}}
\centerline{\includegraphics[width=\linewidth]{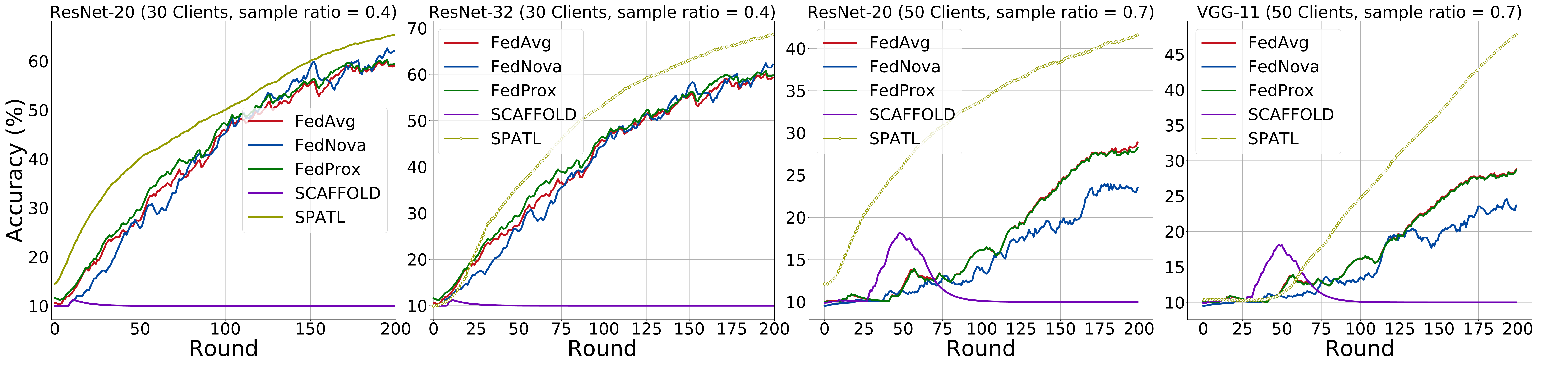}}
\caption{Comparison of SPATL with SoTAs: the top-1 average test accuracy vs. communication rounds. The VGG-11~\cite{simonyan2015vgg} and ResNet-20/32~\cite{he2016resnet} are trained on CIFAR-10~\cite{krizhevsky2009cifar}. 2-layer CNN is trained on FEMNIST~\cite{caldas2019leaf}.}
\vspace{-1em}
\label{fig:vgg_cifar}
\end{center}
 \end{figure*}

\subsubsection{Aggregation with Salient Parameters}
Due to the non-IID local training data in heterogeneous clients, salient parameter selection policy varies among the heterogeneous clients after local updates. Since the selected salient parameters have different matrix sizes and/or dimensions, directly aggregating them will cause a matrix dimension mismatch. To prevent this, as Figure~\ref{fig:pram_aggre} shows, we only aggregate partial parameters according to the current client's salient parameter index on the server side. 
Equation \ref{eq:aggre} shows the mathematical representation of this process.
\begin{equation}
\label{eq:aggre}
    w_g \leftarrow  w_g + \eta(\frac{1}{|K|} \sum_{k \in K} (w_g[i^k,:,:] - w^k)) 
\end{equation}
Here, $w_g$ is the global parameter, $w_k$ is the $i^{th}$ client's salient parameter, $i^k$ is the $w^k$'s index corresponding to the original weights, and $\eta$ is the update step size. 
By only aggregating the salient parameter $w^k$ and its corresponding index $i^k$~(negligible burdens), we can significantly reduce the communication overhead and avoid matrix dimension mismatches.

\section{Experiment}
\label{sec:eval}
We conducted extensive experiments to examine \proj's performance. Overall, we divided our experiments into three categories: learning efficiency, communication cost, and inference acceleration. We also performed an ablation study and compared \proj with state-of-the-art FL algorithms.
\begin{figure}
 \begin{center}

\centerline{\includegraphics[width=\linewidth]{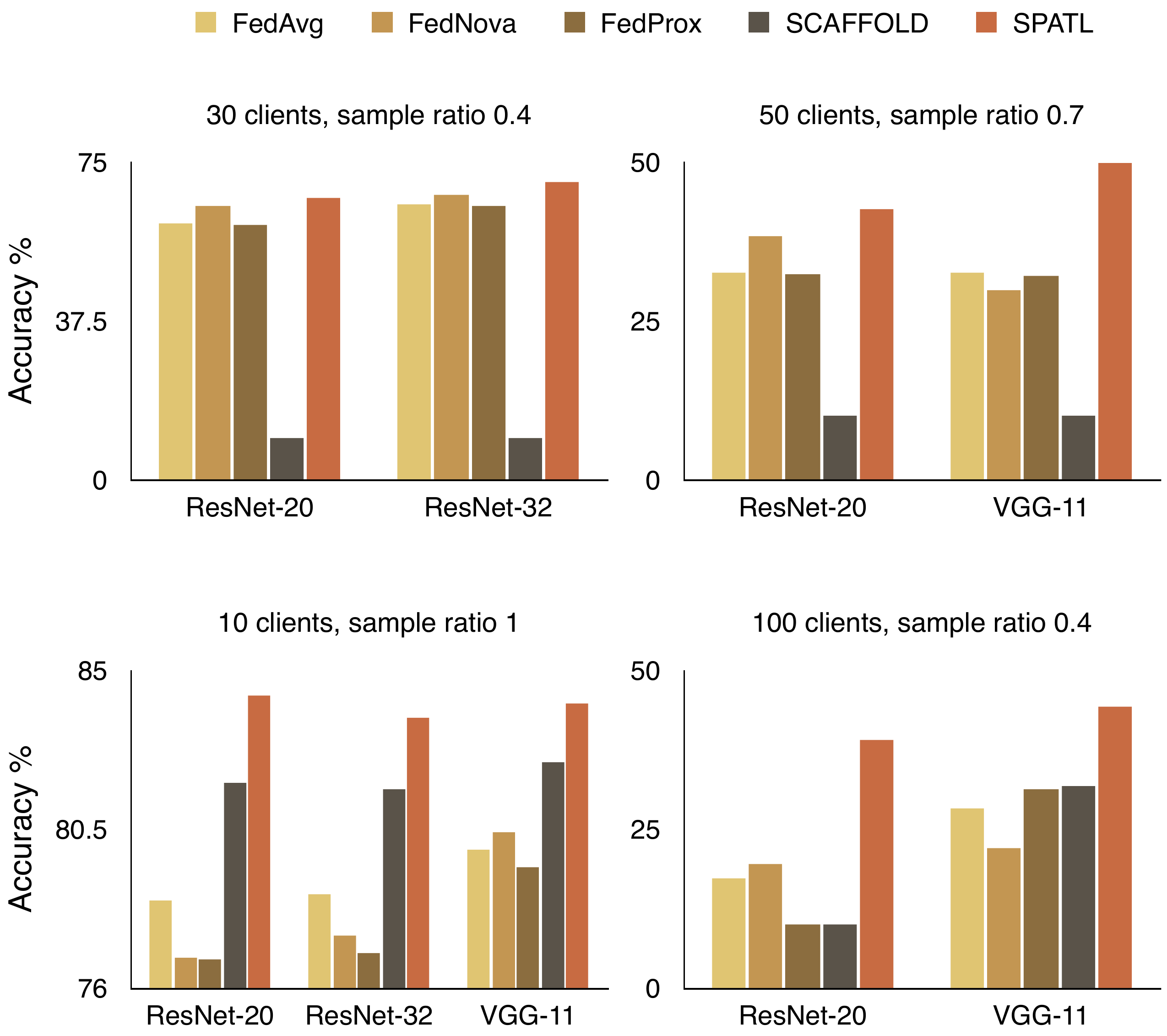}}
\caption{Final accuracy at convergence, the higher the better.}
\vspace{-1em}
\label{fig:conv_acc}
\end{center}
 \end{figure}


\subsection{Implementation and Hyper-parameter Setting}


\textbf{Datasets and Models.}
The experiments are conducted with FEMNIST~\cite{caldas2019leaf} and CIFAR-10~\cite{krizhevsky2009cifar}. In FEMNIST, we follow the LEAF benchmark federated learning setting~\cite{caldas2019leaf}. In CIFAR-10, we use the Non-IID benchmark federated learning setting~\cite{li2022NonIIDBench}. Each client is allocated a proportion of the samples of each label according to Dirichlet distribution (with concentration $\alpha$). Specifically, we sample $p_k \sim Dir_N(\alpha)$ and allocate a $p_{k,j}$ proportion of the instances to client $j$. Here we choose the $\alpha = 0.1$.
The deep learning models we use in the experiment are VGG-11~\cite{simonyan2015vgg} and ResNet-20/32~\cite{he2016resnet}.


\textbf{Federated Learning Setting.}
We follow the Non-IID benchmark federated learning setting and implementation ~\cite{li2022NonIIDBench}. 
In \proj, the models in each client are different. Thus, we evaluate the average performance of models in heterogeneous clients.
We experiment on different clients and sample ratio~(percentage of participating clients in each round) setting, from 10 clients to 100 clients, and the sample ratio from 0.4 to 1. 
During local updates, each client updates 10 rounds locally. The detailed setting can be found in supplementary materials.

\textbf{RL Agent Settings.}
The RL agent is pre-trained on ResNet-56 by a network pruning task. Fine-tuning the RL agent in the first 10 communication rounds with 20 epochs in each updating round. We only update the MLP's (i.e., output layers of RL policy network) parameter when fine-tuning. We use the PPO~\cite{schulman2017ppo} RL policy, the discount factor is $\gamma = 0.99$, the clip parameter is $0.2$, and the standard deviation of actions is $0.5$. Adam optimizer is applied to update the RL agent, where the learning rate is $3~\times 10^{-4}$ and the $\beta = (0.9, 0.999)$.

\textbf{Baseline.}
We compare \proj with the state-of-the-art FL algorithms, such as FedNova~\cite{wang2020fednova}, FedAvg~\cite{mcmahan2017fedavg}, FedProx~\cite{li2020fedprox}, and SCAFFOLD~\cite{karimireddy2020scaffold}.

\textbf{Experimental Setup.}
Our experimental setup ranges from 10 clients to 100 clients, and our experimental scale is on par with existing state-of-the-art works.
To better compare and fully investigate the optimization ability, some of our experiments~(e.g., communication efficiency) scales are set to be larger than many SOTA method experiments~(such as FedNova~\cite{wang2020fednova}, FedAvg~\cite{mcmahan2017fedavg}, FedProx~\cite{li2020fedprox}, and SCAFFOLD~\cite{karimireddy2020scaffold}).
Recent FL works, such as FedAT~\cite{chai2021fedat}, pFedHN~\cite{Shamsian2021pFedHN}, QuPeD~\cite{ozkara2021quped}, FedEMA~\cite{zhuang2022fedema}, and FedGen~\cite{zhu2021fedgen}, their evaluations are within the same scale as our experiments.
Since the FL is an optimization algorithm, we mainly investigate the training stability and robustness. 
The larger experiment scale will show a similar trend.

\textbf{FL Benchmark.}
We use two standard FL benchmark settings: LEAF~\cite{caldas2019leaf} and Non-IID benchmark~\cite{li2022NonIIDBench}.
LEAF~\cite{caldas2019leaf} provides benchmark settings for learning in FL, with applications including federated learning, multi-task learning, meta-learning, and on-device learning. We use the LEAF to split the FEMNIST into Non-IID distributions.
Non-IID benchmark~\cite{li2022NonIIDBench} is an
experimental benchmark that provides us with Non-IID splitting of CIFAR-10 and standard implementation of SOTAs. Our implementation of FedAvg, FedProx, SCAFFOLD, and FedNova are based on the Non-IID benchmark.

\subsection{Learning Efficiency}
In this section, we evaluate the learning efficiency of \proj by investigating the relationship between communication rounds and the average accuracy of the model.
Since \proj learns a shared encoder, each local client has a heterogeneous predictor, and the model's performance is different among clients. Instead of evaluating a global test accuracy on the server side, we allocate each client a local non-IID training dataset and a validation dataset to evaluate the top-1 accuracy, i.e., the highest probability prediction must be exactly the expected answer, of the model among heterogeneous clients. 
We train VGG-11~\cite{simonyan2015vgg} and ResNet-20/32~\cite{he2016resnet} on CIFAR-10~\cite{krizhevsky2009cifar}, and 2-layer CNN on FEMNIST~\cite{caldas2019leaf} separately until the models converge. We then compare model performance results of \proj with state-of-the-arts~(SoTAs)~(i.e.,  FedNova~\cite{wang2020fednova}, FedAvg~\cite{mcmahan2017fedavg}, FedProx~\cite{li2020fedprox}, and SCAFFOLD~\cite{karimireddy2020scaffold}).

Figure~\ref{fig:vgg_cifar} experiments show 10 clients setting where we sample all 10 clients for aggregation. The effect of heterogeneity is not significant compared to a real-world scale. \proj moderately outperforms the SoTAs on CIFAR-10. Results on the 2-layer CNN model trained on FEMNIST however, is an exception; in this case the model trained by \proj has a slightly lower accuracy than SoTAs. We suspect that it has to do with the small size of the 2-layer CNN and the large data quantity. Particularly, in this case, our ``model over-parameterization" assumption no longer holds, making it hard for the salient parameter selection to fit the training data. To verify our analysis, we increase the complexity of our experiments and conduct further experiments on larger scale FL settings. We increase the number of clients to 30, 50, and 100 with different sample ratios.

As heterogeneity rises with the increase in number of clients, \proj demonstrates superiority in coping with data heterogeneity.
Experiment results in Figure~\ref{fig:vgg_cifar} show that for more complex FL settings, \proj outperforms SoTAs with larger margins. 
In the 30 clients FL setting\footnote{In Fig.~\ref{fig:vgg_cifar} SCAFFOLD\cite{karimireddy2020scaffold} diverges with gradient explosion in Non-IID benchmark settings~\cite{li2022NonIIDBench} when there are more than 10 clients.}, 
for ResNet-20, ResNet-32, and VGG-11,  SPTAL outperforms the SoTA FL methods. Notably, \proj yields a better convergence accuracy and a substantially more stable training process.
In the 50 clients and 100 clients settings, the experiment improvements become more significant, as \proj outperforms the SoTAs by a larger margin. 
Moreover, we noticed that the gradient control based method SCAFFOLD~\cite{karimireddy2020scaffold} suffers from gradient explosion issues when the number of clients increases. Even though we set a tiny learning rate, the explosion problem persists.
Other researchers are facing the same issues when reproducing SCAFFOLD, and our results satisfy finding 6 in ~\cite{li2022NonIIDBench}.



Intuitively, we investigate the model's accuracy overhead. Figure~\ref{fig:conv_acc} shows the converge accuracy comparison with SoTAs. \proj surpasses SoTAs in all the FL settings and achieves higher converge accuracy than SoTAs. Again, the superiority of \proj grows progressively with the heterogeneity of FL settings. For instance, in ResNet-20 with 30 clients, \proj outperforms SoTAs only in terms of final convergence accuracy. However, when we increase to 50 heterogeneous clients, \proj achieves 42.54 \% final accuracy, that is around 10\% higher than FedAvg and FedProx~(they achieve 32.71\% and 32.43\% accuracy, respectively). Additionally, it is worth mentioning that, in the 100 clients experiment setting, we compare the accuracy within 200 rounds since all the baselines diverge in 200 rounds except \proj. This further demonstrates that \proj optimizes and improves the quality of the model progressively and stably.


\begin{figure}
 \begin{center}
\centerline{\includegraphics[width=0.8\columnwidth]{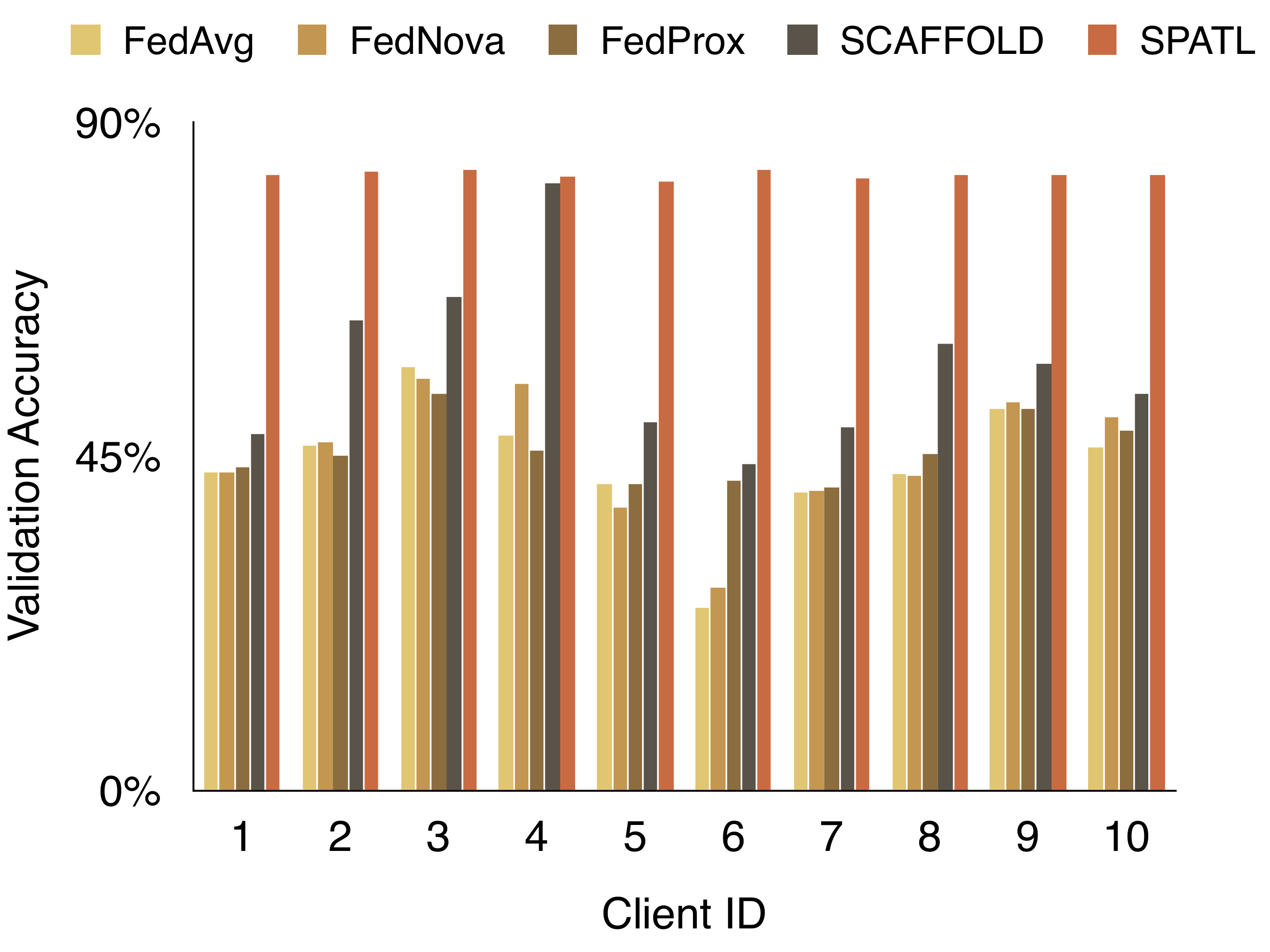}}
\caption{Local validation accuracy for the clients.}
\vspace{-1.5em}
\label{fig:local_acc}
\end{center}
 \end{figure}
Trained model performance on heterogeneous local clients is also an essential indicator when evaluating FL algorithms with regards to deploying AI models on the edge.  
Since edge devices have various application scenarios and heterogeneous input data, models will likely exhibit divergence on such devices.
We further evaluate the robustness and feasibility of FL methods on distributed AI by testing local model accuracies on all clients. 
Figure~\ref{fig:local_acc} shows ResNet-20 on each client's accuracy after the training is complete for CIFAR-10~(total 10 clients trained by \proj and SCAFFOLD with 100 rounds). The model trained by \proj produces better performance across all clients. In particular, the edge model trained by \proj produces more stable performance among each client, whereas models trained by baselines exhibit more variance. For instance, all the edge models trained by \proj have similar accuracy performance. 
Since \proj uses heterogeneous predictors to transfer the encoder's knowledge, the model is more robust when dealing with non-IID data. However, our baseline methods~(such as SCAFFOLD~\cite{karimireddy2020scaffold}) share the entire model when training on non-IID clients, leading to a variance in model performance on non-IID clients and causing poor performance on some clients.



\subsection{Communication Efficiency}
\begin{table*}[]
\centering
\caption{Comparison of communication cost with SoTAs to achieve 80\% accuracy.}
\label{tab:com_cost}
\resizebox{\textwidth}{!}{%
\begin{tabular}{ccccccccc}
\hline
\multirow{2}{*}{\textbf{Method}} & \multirow{2}{*}{\textbf{Model}} & \multirow{2}{*}{\textbf{\begin{tabular}[c]{@{}c@{}}Target\\ Accuracy\end{tabular}}} & \multirow{2}{*}{\textbf{Clients}} & \multirow{2}{*}{\textbf{Communication Rounds}} & \multicolumn{4}{c}{\textbf{Communication Cost}}                                     \\
                                 &                                 &                                                                                     &                                   &                                                & \textbf{Round/Client} & \textbf{Total} & \textbf{$\Delta$ Cost} & \textbf{Speed Up} \\ \hline
\multirow{3}{*}{FedAvg~\cite{mcmahan2017fedavg}}          & ResNet-20                       & \multirow{3}{*}{80\%}                                                               & \multirow{3}{*}{10}               & 203                                            & 2.1MB                 & 4.16GB         & 0GB                    & (1 $\times$)      \\
                                 & ResNet-32                       &                                                                                     &                                   & 192                                            & 3.2MB                 & 6.00GB         & 0GB                    & (1 $\times$)      \\
                                 & VGG-11                          &                                                                                     &                                   & 181                                            & 42MB                  & 74.24GB        & 0GB                    & (1 $\times$)      \\ \hline
\multirow{3}{*}{FedNova~\cite{wang2020fednova}}         & ResNet-20                       & \multirow{3}{*}{80\%}                                                               & \multirow{3}{*}{10}               & 198                                            & 4.2MB                 & 8.12GB         & +3.96GB                & (0.51 $\times$)   \\
                                 & ResNet-32                       &                                                                                     &                                   & 197                                            & 6.4MB                 & 12.31GB        & +6.31GB                & (0.48 $\times$)   \\
                                 & VGG-11                          &                                                                                     &                                   & 159                                            & 84MB                  & 130.43GB       & +56.19                 & (0.57 $\times$)   \\ \hline
\multirow{3}{*}{FedProx~\cite{li2020fedprox}}         & ResNet-20                       & \multirow{3}{*}{80\%}                                                               & \multirow{3}{*}{10}               & 288                                            & 2.1MB                 & 5.91GB         & +1.75GB                & (0.70 $\times$)   \\
                                 & ResNet-32                       &                                                                                     &                                   & 400                                            & 3.2MB                 & 12.80GB        & +6.80GB                & (0.47 $\times$)   \\
                                 & VGG-11                          &                                                                                     &                                   & 296                                            & 42MB                  & 121.41GB       & +47.17GB               & (0.61 $\times$)   \\ \hline
\multirow{3}{*}{SCAFFOLD~\cite{karimireddy2020scaffold}}        & ResNet-20                       & \multirow{3}{*}{80\%}                                                               & \multirow{3}{*}{10}               & 49                                             & 4.3MB                 & 2.10GB         & -2.06GB                & (1.98 $\times$)   \\
                                 & ResNet-32                       &                                                                                     &                                   & 55                                             & 6.5MB                 & 3.60GB         & -2.40GB                & (1.67 $\times$)   \\
                                 & VGG-11                          &                                                                                     &                                   & 54                                             & 84MB                  & 45.00GB        & -29.24GB               & (1.65 $\times$)   \\ \hline
\multirow{3}{*}{SPATL (Ours)}    & ResNet-20                       & \multirow{3}{*}{80\%}                                                               & \multirow{3}{*}{10}               & 52                                             & 2.1MB                 & 1.10GB         & -3.06GB                & (3.78 $\times$)   \\
                                 & ResNet-32                       &                                                                                     &                                   & 50                                             & 3.6MB                 & 1.80GB         & -4.20GB                & (3.33 $\times$)   \\
                                 & VGG-11                          &                                                                                     &                                   & 46                                             & 61.3MB                & 28.00GB        & -46.24GB               & (2.65 $\times$)   \\ \hline
\end{tabular}%
}
\end{table*}
A key contribution of \proj that makes it stand out among SoTAs is its significant reduction of communication overhead due to salient parameter selection. Although
SoTAs, like FedNova~\cite{wang2020fednova} and SCAFFOLD~\cite{karimireddy2020scaffold}, achieve stable training via gradient control or gradient normalization variates, their average communication cost doubles compared to FedAvg~\cite{mcmahan2017fedavg} as a result of sharing the extra gradient information.
We present two experiment settings to evaluate model communication efficiencies. First, we trained all models to a target accuracy and calculated communication cost. Second, we trained all models to converge and calculated the communication cost of each FL algorithm.
The communication cost is calculated as:
\begin{equation}
    \# \text{Rounds} \times \text{Client's round cost} \times \# \text{Sampled Clients}
\end{equation}

\begin{figure}
 \begin{center}

\centerline{\includegraphics[width=\linewidth]{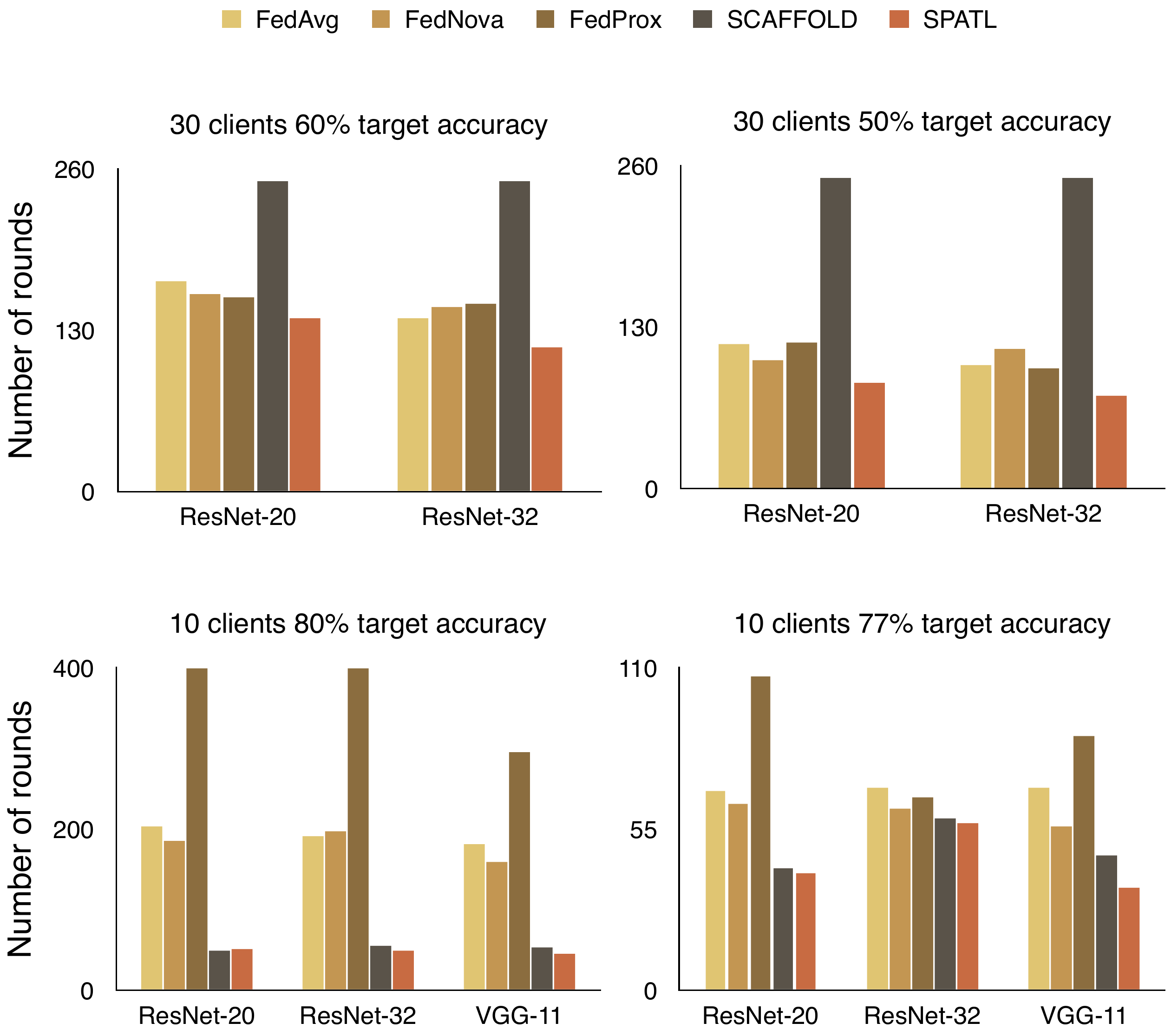}}
\caption{Number of rounds required by different models to reach the target accuracy, the lower the better.}
\vspace{-1em}
\label{fig:train_rounds}
\end{center}
 \end{figure}

Table~\ref{tab:com_cost} shows the detailed information of communication cost~(FedAvg~\cite{mcmahan2017fedavg} as benchmark) when we train models to a target accuracy.
\proj remarkably outperforms SoTAs. In ResNet-20, \proj reduced communication cost by up to $7.6\times$~(FedNova costs 8.12GB while \proj only costs 1.1GB). Moreover, \proj reduced communication by up to 102GB when training VGG-11 compared to FedNova.

There are two main benefits of \proj in reducing communication overhead. 
First, training in \proj is much more efficient and stable. Our experiments show that \proj requires fewer communication rounds to achieve target accuracy. 
For example, in VGG-11, FedProx uses 296 training rounds while \proj uses 250 less rounds to achieve the same target accuracy. This significantly reduces the communication cost. 
We provide a more comprehensive comparison to show the number of rounds different models take to achieve target accuracy. As Figure~\ref{fig:train_rounds} shows, we try different FL settings, and \proj consistently requires fewer rounds than SoTAs in most of the FL settings~(except in ResNet-20 with 10 clients and 80\% target accuracy, SPATL requires 3 rounds more than SCAFFOLD. However, as shown in Table~\ref{tab:com_cost}), the total communication cost of \proj is significant less than all others.
Second, since the salient parameter selection agent selectively uploads partial parameters, \proj significantly reduces communication cost. As shown in Table~\ref{tab:com_cost}, compared to the gradient control based methods, such as SCAFFOLD~\cite{karimireddy2020scaffold} and FedNova~\cite{wang2020fednova}, \proj remarkably reduces per round communication cost. For instance, \proj uses $2 \times$ less round costs in ResNet-20 compared to the traditional FL, such as FedAvg~\cite{mcmahan2017fedavg}. Even when factoring in gradient control information, salient parameter selection enables \proj to drop unnecessary communication burdens, which makes its round costs comparable to FedAvg.

Furthermore, we investigated the convergence accuracy of models using SoTAs and \proj optimizations.
We consider a performance upper bound by creating a hypothetical centralized case where images are heterogeneously distributed across 30, 50, and 100 clients.
Table~\ref{tab:com_cost2} shows the results of training the models to convergence. Compared to FedAvg~\cite{mcmahan2017fedavg}, gradient control based FL algorithms have higher accuracy at the expense of communication efficiency. For instance, FedNova~\cite{wang2020fednova} achieves slightly higher accuracy. However, their communication budget increases by more than $2 \times$. Models optimized by \proj achieve significantly higher accuracy than all other baselines. Especially on VGG-11 with 50 clients, \proj achieves 17.8\%, 19.86\%, and 17.62\% higher accuracy than FedAvg, FedNova, and FedProx respectively.  
Particularly, unlike FedNova, which sacrifices communication cost for higher accuracy, \proj takes advantage of salient parameter selection to achieve higher accuracy with relatively negligible communication overhead.
For instance, when training the ResNet-20 for 30 and 50 clients, \proj achieves the best accuracy with the lowest communication cost. The results further show that \proj remarkably reduces the communication cost and has a higher capacity to train models for better performance.


\begin{table*}[]
\centering
\caption{Comparison of communication cost with SoTAs to achieve model convergence.}
\label{tab:com_cost2}
\resizebox{\textwidth}{!}{%
\begin{tabular}{cccccccccc}
\hline
\multirow{2}{*}{\textbf{Method}} & \multirow{2}{*}{\textbf{Model}} & \multirow{2}{*}{\textbf{Clients}} & \multirow{2}{*}{\textbf{\begin{tabular}[c]{@{}c@{}}Sample \\ Ratio\end{tabular}}} & \multirow{2}{*}{\textbf{Converge Rounds}} & \multicolumn{3}{c}{\textbf{Communication Cost}}             & \multirow{2}{*}{\textbf{\begin{tabular}[c]{@{}c@{}}Avg.\\ Converge Acc.\end{tabular}}} & \multirow{2}{*}{\textbf{\begin{tabular}[c]{@{}c@{}}$\Delta$ \\ Acc.\end{tabular}}} \\
                                 &                                 &                                   &                                                                                   &                                           & \textbf{Round/Client}   & \textbf{Total} & \textbf{Speedup} &                                                                                        &                                                                                    \\ \hline
\multirow{5}{*}{FedAvg~\cite{mcmahan2017fedavg}}          & \multirow{2}{*}{ResNet-20}      & 30                                & 0.4                                                                               & 170                                       & \multirow{2}{*}{2.1MB}  & 4.18GB         & (1 $\times$)     & 60.58\%                                                                                & 0\%                                                                                \\
                                 &                                 & 50                                & 0.7                                                                               & 184                                       &                         & 13.21GB        & (1 $\times$)     & 32.71\%                                                                                & 0\%                                                                                \\
                                 & ResNet-32                       & 30                                & 0.4                                                                               & 140                                       & 3.2MB                   & 5.25GB         & (1 $\times$)     & 64.92\%                                                                                & 0\%                                                                                \\
                                 & \multirow{2}{*}{VGG-11}         & 50                                & 0.7                                                                               & 171                                       & \multirow{2}{*}{42MB}   & 246GB          & (1 $\times$)     & 32.68\%                                                                                & 0\%                                                                                \\
                                 &                                 & 100                               & 0.4                                                                               & 322                                          &                         &          528GB      & (1 $\times$)     &      28.33\%                                                                                  & 0\%                                                                                \\ \hline
\multirow{5}{*}{FedNova~\cite{wang2020fednova}}         & \multirow{2}{*}{ResNet-20}      & 30                                & 0.4                                                                               & 159                                       & \multirow{2}{*}{4.2MB}  & 7.83GB         & (0.54 $\times$)  & 64.50\%                                                                                & +3.92\%                                                                            \\
                                 &                                 & 50                                & 0.7                                                                               & 200                                       &                         & 28.71GB        & (0.46 $\times$)  & 38.35\%                                                                                & +5.64\%                                                                            \\
                                 & ResNet-32                       & 30                                & 0.4                                                                               & 149                                       & 6.4MB                   & 11.18GB        & (0.47 $\times$)  & 67.48\%                                                                                & +2.56\%                                                                            \\
                                 & \multirow{2}{*}{VGG-11}         & 50                                & 0.7                                                                               & 193                                       & \multirow{2}{*}{84MB}   & 554GB          & (0.44 $\times$)  & 29.90\%                                                                                & -2.78\%                                                                            \\
                                 &                                 & 100                               & 0.4                                                                               & 287                                          &                         &       941GB         & (0.56 $\times$)     &     22.12\%                                                                                   &     -6.21\%                                                                               \\ \hline
\multirow{5}{*}{FedProx~\cite{li2020fedprox}}         & \multirow{2}{*}{ResNet-20}      & 30                                & 0.4                                                                               & 157                                       & \multirow{2}{*}{2.1MB}  & 3.86GB         & (1.08 $\times$)  & 60.20\%                                                                                & -0.38\%                                                                            \\
                                 &                                 & 50                                & 0.7                                                                               & 193                                       &                         & 13.85GB        & (0.95 $\times$)  & 32.43\%                                                                                & -0.28\%                                                                            \\
                                 & ResNet-32                       & 30                                & 0.4                                                                               & 151                                       & 3.2MB                   & 5.66GB         & (0.93 $\times$)  & 64.75\%                                                                                & -0.17\%                                                                            \\
                                 & \multirow{2}{*}{VGG-11}         & 50                                & 0.7                                                                               & 195                                       & \multirow{2}{*}{42MB}   & 280GB          & (0.88 $\times$)  & 32.14\%                                                                                & -0.54\%                                                                            \\
                                 &                                 & 100                               & 0.4                                                                               & 123                                          &                         &        201GB        & (2.63 $\times$)     &      31.21\%                                                                                  &      +2.88\%                                                                              \\ \hline
\multirow{5}{*}{SCAFFOLD~\cite{karimireddy2020scaffold}}        & \multirow{2}{*}{ResNet-20}      & 30                                & 0.4                                                                               & 400                                       & \multirow{2}{*}{4.3MB}  & 20.16GB        & (0.20 $\times$)  & 10.00\%                                                                                   & -50.58\%                                                                           \\
                                 &                                 & 50                                & 0.7                                                                               & 400                                       &                         & 58.79GB        & (0.22 $\times$)  & 10.00\%                                                                                   & -22.71\%                                                                           \\
                                 & ResNet-32                       & 30                                & 0.4                                                                               & 400                                       & 6.5MB                   & 30.47GB        & (0.17 $\times$)  & 10.00\%                                                                                   & -54.92\%                                                                           \\
                                 & \multirow{2}{*}{VGG-11}         & 50                                & 0.7                                                                               & 400                                       & \multirow{2}{*}{84MB}   & 1148GB         & (0.21 $\times$)  & 10.00\%                                                                                   & -22.68\%                                                                           \\
                                 &                                 & 100                               & 0.4                                                                               & 400                                       &                         &      1312GB          & (0.40 $\times$)     & 31.78\%                                                                                   &      +3.45\%                                                                              \\ \hline
\multirow{5}{*}{SPATL (Ours)}    & \multirow{2}{*}{ResNet-20}      & 30                                & 0.4                                                                               & 140                                       & \multirow{2}{*}{2.1MB}  & 3.44GB         & (1.21 $\times$)  & 66.49\%                                                                                & +5.91\%                                                                            \\
                                 &                                 & 50                                & 0.7                                                                               & 171                                       &                         & 12.27GB        & (1.08 $\times$)  & 42.54\%                                                                                & +9.83\%                                                                            \\
                                 & ResNet-32                       & 30                                & 0.4                                                                               & 117                                       & 3.6MB                   & 4.93GB         & (1.06 $\times$)  & 70.23\%                                                                                & +5.31\%                                                                            \\
                                 & \multirow{2}{*}{VGG-11}         & 50                                & 0.7                                                                               & 156                                       & \multirow{2}{*}{61.3MB} & 324GB          & (0.76 $\times$)  & 49.76\%                                                                                & +17.08\%                                                                           \\
                                 &                                 & 100                               & 0.4                                                                               & 86                                          &                         &        205GB        & (2.58 $\times$)     &      44.23\%                                                                                  &     +15.90\%                                                                               \\ \hline
\end{tabular}%
}
\end{table*}


\subsection{Inference Acceleration}
\begin{table}
  \caption{Inference acceleration.}
  \label{tab:inference}
  \centering
  \resizebox{\linewidth}{!}{
  \begin{tabular}{cccc}
    \toprule
    Model  & Avg. sparsity ratio & Avg. FLOPs $\downarrow$  & Highest FLOPs $\downarrow$\\
    \midrule
    VGG-11    & $70\%$  & $29.9\%  $ & $39.7\% $  \\

    ResNet-20    & $84\%$ & $25.8\%$  & $31.9\%  $  \\
    ResNet-32    & $82\%$ & $29.6\%  $ & $38.4\%  $\\
    \bottomrule
  \end{tabular}}

\end{table}
In this section, we evaluate the inference acceleration of \proj. 
Local inference is a crucial measure for deploying AI models on the edge since edge devices have limited computing power, and edge applications~(e.g., self-driving car) are inference sensitive.
In \proj, when the salient parameter selection agent selects salient parameters, it prunes the model as well.
For a fair evaluation of inference, instead of recording the actual run time~(run time may vary on different platforms) of pruned models, we calculated the FLOPs~(floating point operations per second).

Table~\ref{tab:inference} shows the inference acceleration status after training is complete. \proj notably reduced the FLOPs in all the evaluated models. For instance, in ResNet-32, the average FLOPs reduction among 10 clients is $29.6\%$, and the client with the highest FLOPs reduction achieves $38.4\%$ fewer FLOPs than the original model, while the client models have a relatively low sparsity ratio~(the sparsity ratio represents the ratio of salient parameters compared to the entire model parameters). The low sparsity ratio can further benefit by accelerating the models on parallel platforms, such as GPUs. 
Additionally, we evaluate the salient parameter selection agents' pruning ability and compare it with SoTA pruning methods. As shown in Table~\ref{tab:res_cifar}, our agent achieves outstanding results in pruning task and outperforms popular AutoML pruning baselines. The results indicate that \proj can significantly accelerate model inference with acceptably small accuracy loss.


\begin{table}
\caption{Transferability of model trained by FL.}
  \label{tab:trans}
  \centering
\resizebox{0.85\linewidth}{!}{\begin{tabular}{cccc}
\toprule
Method   & Model                  & Org. Acc. &  Transferred  Acc.     \\
\midrule
FedAvg~\cite{mcmahan2017fedavg}   & \multirow{5}{*}{ResNet-20} &   78.5   &         92.16                                  \\
FedNova~\cite{wang2020fednova} &                        &    76.9  &         93.66                                  \\
FedProx~\cite{li2020fedprox}  &                        &   76.8   &        93.03                                    \\
SCAFFOLD~\cite{karimireddy2020scaffold} &                        &    81.8  &      91.7                                     \\
SPATL (Ours)    &                        &     84.3 &       93.83                                     \\
\bottomrule

\end{tabular}
}
\end{table}
\subsection{Transferbility of Learned Model}
In \proj, since only the partial model~(i.e., knowledge encoder) is trained in a distributed manner, we conducted a transferability comparison experiment to test for successful transfer of knowledge among heterogeneous edge clients.
Specifically, we transfer the neural network trained by \proj and SoTAs (e.g., FedAvg, FedNova, SCAFFOLD, etc.) separately to a new portion of data and compare the performance of transferred models.
The experimental settings are as follows:
we split the CIFAR-10~\cite{krizhevsky2009cifar} into two separate datasets, one with 50K images~(for federated learning) and another with 10K images~(for transfer learning after FL is finished). We use ResNet-20 and set 10 clients for federated learning, where each client has 4k image local training data and 1k validation set. Transfer learning was conducted in a regular manner without involving training distribution.

Table~\ref{tab:trans} shows the results. 
The model trained by \proj achieves comparable transfer learning results with the SoTAs.
This further shows that \proj, which only trains a shared encoder in a distributed manner~(i.e., as opposed to training the entire model), can successfully learn and transfer the knowledge of a heterogeneous dataset.

\begin{figure*}
\begin{center}

\centerline{\includegraphics[width=.71\linewidth]{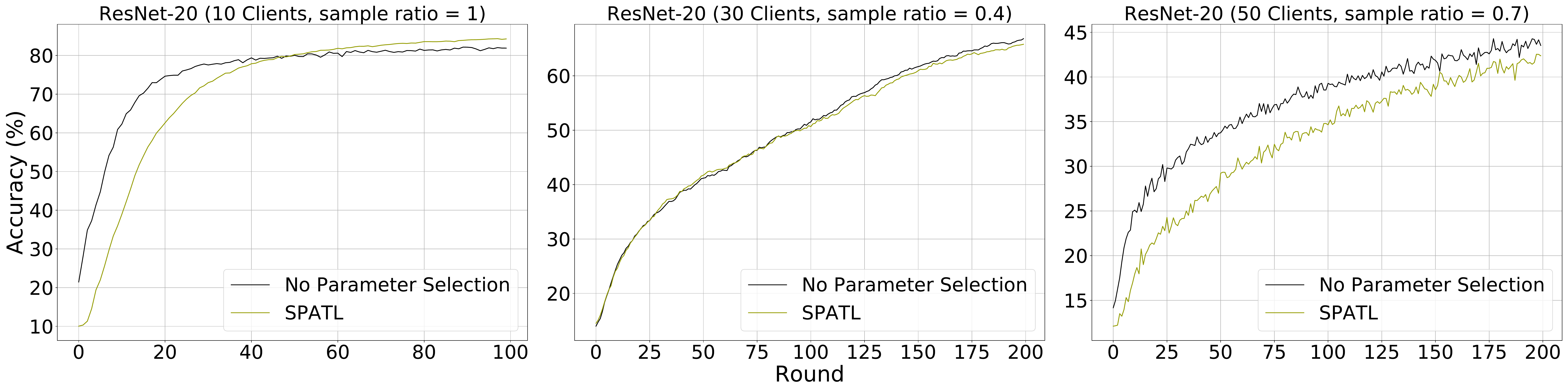}}
\caption{Parameter selection vs. no parameter selection.}

\label{fig:pram_ablation}
\end{center}
\end{figure*}
\begin{figure}%
 \vskip -0.14in

    \centering
    \subfloat[\centering]{{\includegraphics[width=.49\linewidth]{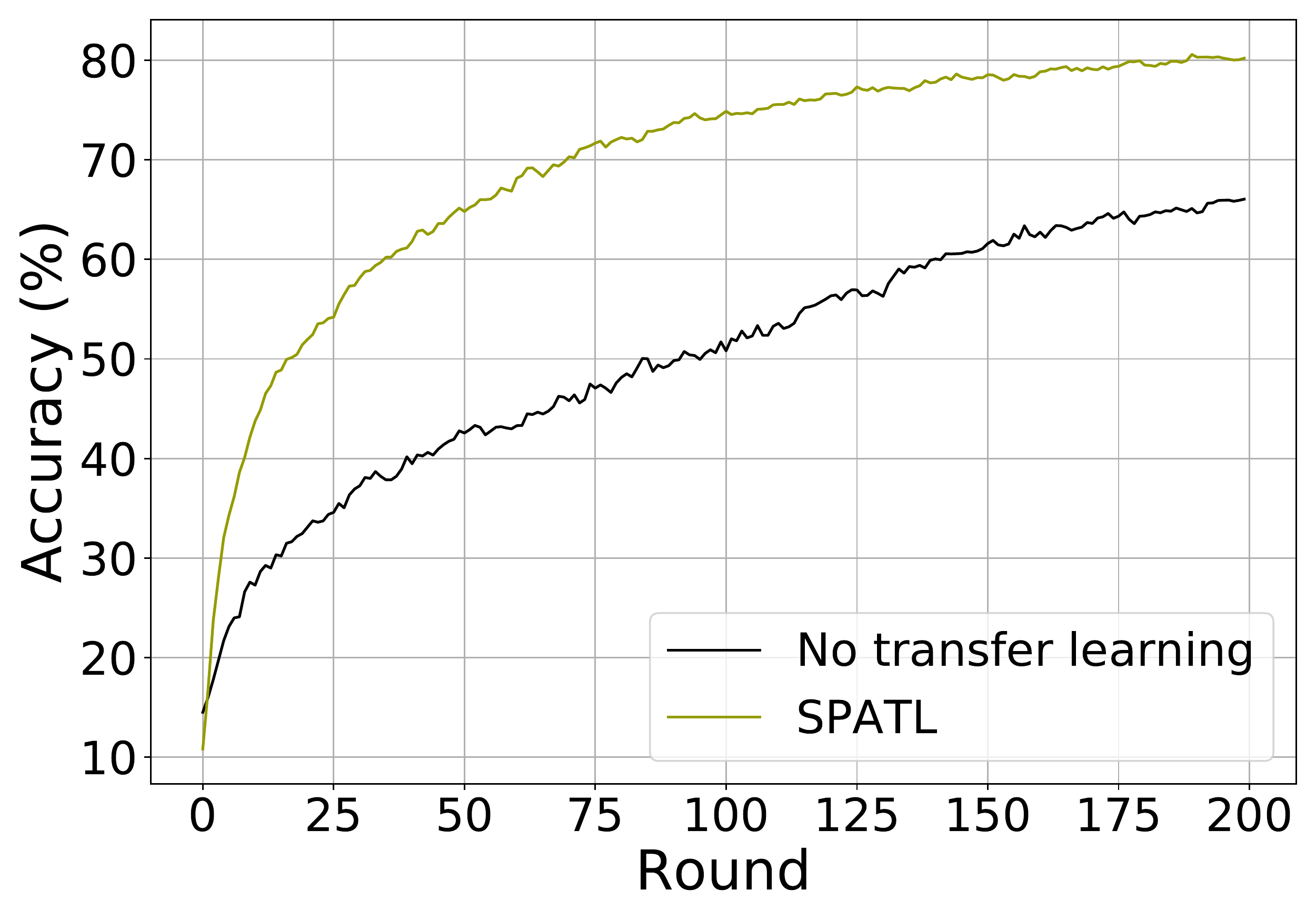} }}%
    \subfloat[\centering]{{\includegraphics[width=.49\linewidth]{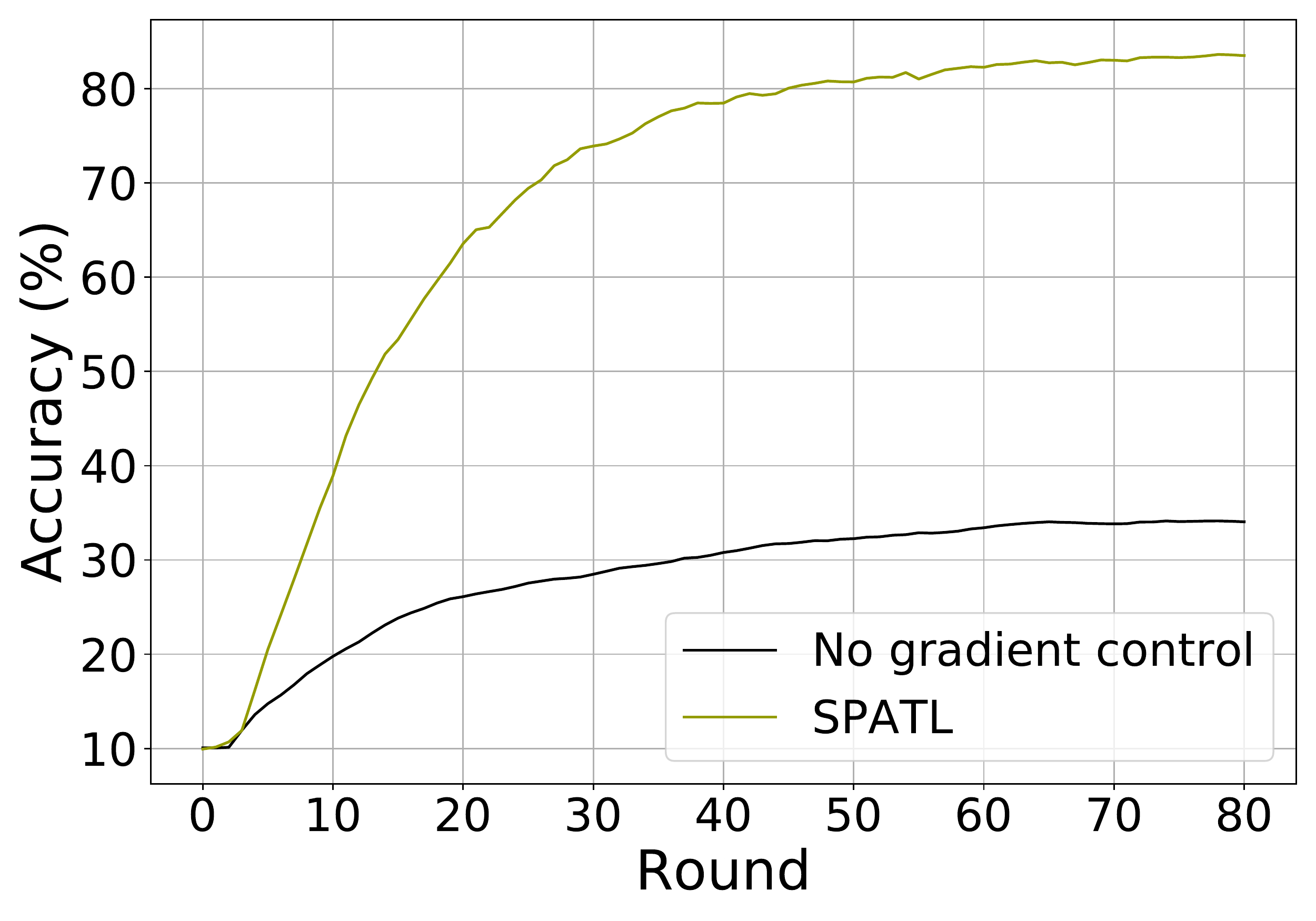} }}%
    \caption{(a) Transfer learning vs. no transfer learning. (b) Gradient control vs. no gradient control.}%
    \label{fig:ablation}%
 \vskip -0.2in
\end{figure}
\subsection{Ablation Study}
\subsubsection{Salient Parameter Selection vs. No Parameter Selection}
Modern AI models are huge (involving billions of parameters) and often over-parameterized. 
Thus, only a subset of salient parameters can significantly affect the model's final performance. As such, a reasonable pruning of redundant parameters might not negatively impact model training. This section investigates the impact of salient parameter selection on federated learning. Specifically, we compare \proj with and without salient parameter selection.
\begin{table}[t]
\caption{Pruning comparison: top-1 accuracy results for pruned ResNets.}
\label{tab:res_cifar}
\vspace{-10pt}
\begin{center}
\begin{small}

\resizebox{0.85\linewidth}{!}{\begin{tabular}{lccccccr}
\toprule
Model   & Method & FLOPs $\downarrow$  &Top-1 & $\Delta $Top-1 \\

\midrule
\multirow{5}{*}{ResNet-20}    
                          &Uniform& $50\%$ &  $84.00$ & $-7.73$ \\
                          &AMC~\cite{he2018amc}    & $50\%$ &  $86.40$ & $-5.33$ \\
                          &AGMC~\cite{yu2021agmc}   & $50\%$ &  $88.42$ & $-3.31$ \\
                          &SFP~\cite{he2018sfp}   & $42\%$ &  $90.83$ & $-1.37$ \\
                          &FPGM~\cite{he2019FPGM}   & $42\%$ &  $91.09$ & $-1.11$ \\
                          &DSA~\cite{ning2020dsa}   & $50\%$ &  $91.38$ & $-0.79$ \\
                          &{SPATL (Ours)} & ${51\%}$   &  ${91.31}$ & ${-0.42}$ \\

\bottomrule
\end{tabular}
}

\end{small}

\end{center}
\vspace{-15pt}
\end{table}

Figure~\ref{fig:pram_ablation} shows the results.
We conducted the experiment on ResNet-20 with various FL settings. All of the results indicate that 
properly pruning some unimportant weights of over-parameterized networks will not harm training stability in federated learning. Instead, it might produce better results in some cases. Especially in the 10 clients setting, \proj optimized a higher quality model after applying the parameter selection. 
We further evaluate the salient parameter agent by applying it to network pruning task and comparing it with popular pruning methods.
Table~\ref{tab:res_cifar} shows the pruning results. \proj's parameter selection agent achieves competitive results with SoTA pruning methods, which means that it can prune redundant parameters and significantly reduce the FLOPs or pruned model with negligible accuracy loss. 
Moreover, state-of-the-art salient parameter selection methods, such as SFP~\cite{he2018sfp}, DSA~\cite{ning2020dsa}, and FPGM~\cite{he2019FPGM}, are usually non-transferable for a given model. They require time-consuming search and re-training to find a target model architecture and salient parameters. For instance, as shown in~\cite{wang2020apq}~(table 2), a combined model compression method needs 85.08 lbs of $CO_{2}$  emissions to find a target model architecture. This makes it expensive to deploy on edge devices. 
In \proj, the RL agent is a tiny GNN followed by an MLP. The cost to compute target salient parameters within one-shot inference~(0.36 ms on NVIDIA V100) and the memory consumption is 26 KB, which is acceptable on edge devices.

\subsubsection{Transfer Learning vs. No Transfer Learning}
\proj transfers the shared encoder to local non-IID datasets and addresses the heterogeneous issue of FL. To investigate the effects of transfer learning on \proj, in this section, we disable \proj's transfer learning. Figure~\ref{fig:ablation} (a) shows the results. We train the ResNet-20~\cite{he2016resnet} on CIFAR-10~\cite{krizhevsky2009cifar} with 10 clients and sample all the clients in communication. \proj without transfer learning has a poor performance when optimizing the model. Combining the results present in Figure~\ref{fig:local_acc}, we can infer that a uniform model deployed on heterogeneous clients can cause performance diversity~(i.e., the model performs well on some clients but poor on others). Intuitively, clients with data distribution similar to global data distribution usually perform better; nevertheless, clients far away from global data distribution are hard to converge.
It is adequate to show that by introducing transfer learning, \proj can better deal with heterogeneous issues in FL. 
Transfer learning enables every client to customize the model on its non-IID data and produces significantly better performance than without transfer learning.

\subsubsection{Impact of Gradient Control}
\proj maintains control variates both in the local and cloud environment to help correct the local update directions and guide the encoder's local gradient towards the global gradient direction.
Figure~\ref{fig:ablation} (b) shows the results of \proj with and without gradient control. We train VGG-11~\cite{simonyan2015vgg} on CIFAR-10~\cite{krizhevsky2009cifar} with 10 clients. 
Training the model in the heterogeneous non-IID local dataset typically causes high variants of local gradients leading to poor convergence. The gradient control variates in \proj maintain the global gradient direction and correct the gradient drift, thus producing a stable training process. 
The results are in line with our expectations that gradient control remarkably improves the training performance of \proj. 
\begin{figure}
 \begin{center}
    \centering

\centerline{\includegraphics[width=.49\columnwidth]{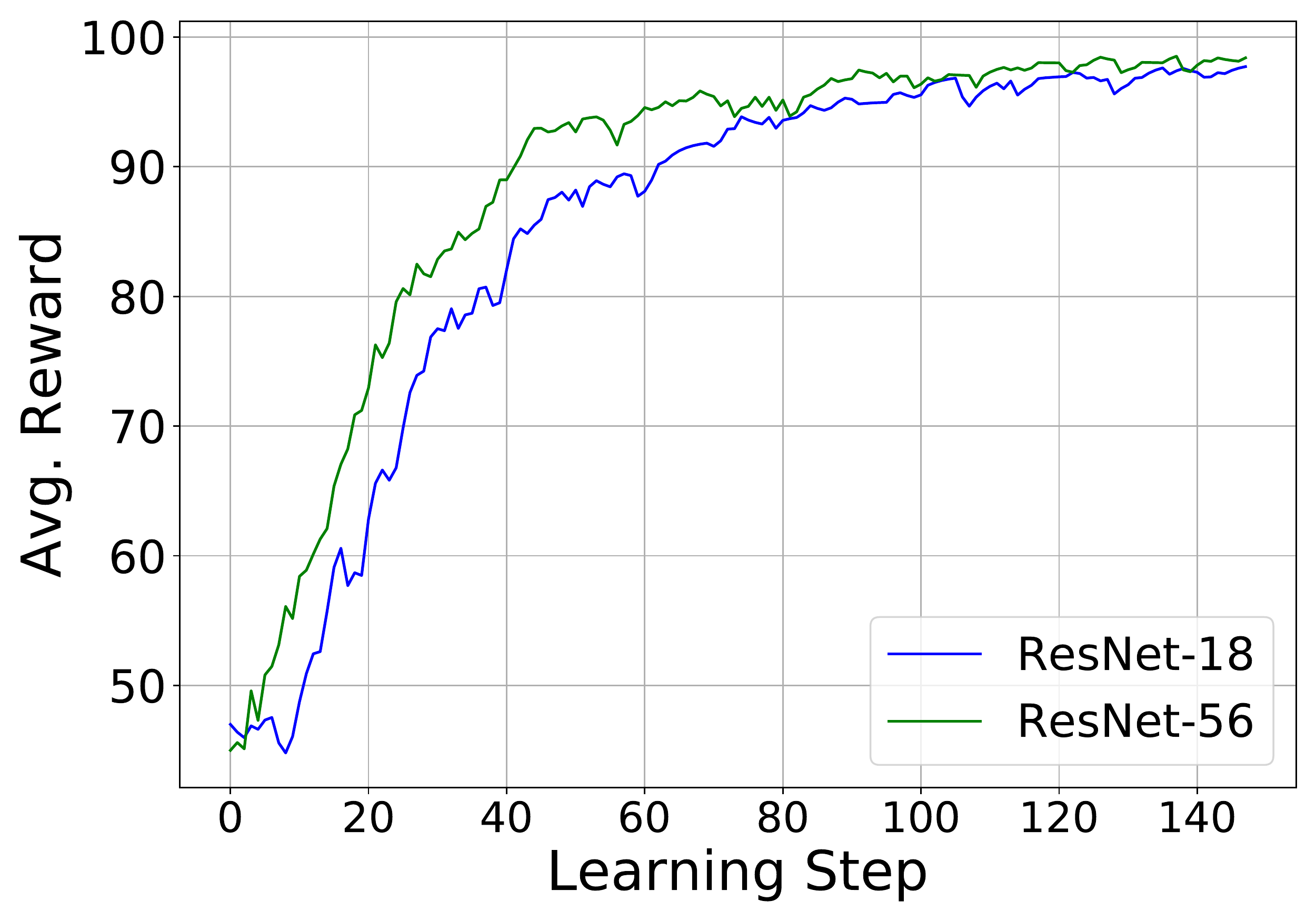}}
\caption{RL's average reward vs. update round.}
\label{fig:rl_ef}
\end{center}
 \end{figure}
\subsubsection{Fine-tuning Reinforcement Learning Agent}
This section discusses the cost of pre-training a reinforcement learning agent and the cost of customization by slightly fine-tuning the agent's weights through online reinforcement learning.
We pre-train the RL-agent and perform network pruning on ResNet-56, then transfer the agent to ResNet-18 by fine-tuning, and only update the predictor of RL-agent's policy network. 
Figure~\ref{fig:rl_ef} shows the average reward the RL agent gets on network pruning task and the agent's corresponding update round. In both ResNet-18 and ResNet-56, the RL agent converges rapidly around 40 rounds of RL policy updating. Particularly, in ResNet-18, by slightly fine-tuning the RL-agent, it achieves comparable rewards to ResNet-56. That means the agent can be successfully transferred to a newly deployed model. This further shows the feasibility of fine-tuning the pre-trained salient parameter selection agent.  



\section{Conclusion and Discussion}
\label{sec:con}

 In this paper, we presented \proj, a method for efficient federated learning using salient parameter aggregation and transfer learning. To address data heterogeneity in federated learning, we introduced a knowledge transfer local predictor that transfers the shared encoder to each client. We proposed a salient parameter selection agent to filter salient parameters of the over-parameterized model before communicating it with the server. As a result, the proposed method significantly decreases the communication overhead. We further leveraged a gradient control mechanism to stabilize the training process and make it more robust. Our experiments show that \proj has a stable training process and achieves promising results. Moreover, \proj significantly reduces the communication cost and accelerates the model inference time.
 The proposed approach may have poor performance on simple models. As Figure~\ref{fig:vgg_cifar} shows, our approach works well on over-parameterized neural networks, such as ResNet~\cite{he2016resnet} and VGG~\cite{simonyan2015vgg} net. However, when it turns to less-parameterized models, such as 2-layer CNNs, the salient parameter selection may degrade in performance, making the model converge slower than baselines. In practice, less-parameterized models are rarely used in real-world applications.
Second, not all AI models are transferable. 
In our future work, we will continuously improve the universality of our method.

\bibliographystyle{IEEEtran}
\bibliography{IEEEabrv}


\end{document}